\documentclass[twocolumn,showpacs,prl]{revtex4-1}
\usepackage{amsmath}
\usepackage{amsfonts}
\usepackage{graphicx}

\usepackage{color}
\usepackage{algorithm2e}
\usepackage{comment}

\DeclareMathOperator{\KL}{KL}
\DeclareMathOperator{\I}{I}
\DeclareMathOperator{\HS}{H}
\DeclareMathOperator{\NTIC}{NTIC}
\DeclareMathOperator{\pred}{pred}
\DeclareMathOperator{\past}{past}
\DeclareMathOperator{\future}{future}

\begin{document}
\title{Neural Coarse-Graining: Extracting slowly-varying latent degrees of freedom with neural networks}
\author{Nicholas Guttenberg}
\affiliation{Araya, Tokyo}
\affiliation{Earth-Life-Science Institute, Tokyo}
\author{Martin Biehl}
\affiliation{University of Hertfordshire, Hatfield}
\author{Ryota Kanai}
\affiliation{Araya, Tokyo}

\begin{abstract}
 We present a loss function for neural networks that encompasses an idea of trivial versus non-trivial predictions, such that the network jointly determines its own prediction goals and learns to satisfy them. This permits the network to choose sub-sets of a problem which are most amenable to its abilities to focus on solving, while discarding 'distracting' elements that interfere with its learning. To do this, the network first transforms the raw data into a higher-level categorical representation, and then trains a predictor from that new time series to its future. To prevent a trivial solution of mapping the signal to zero, we introduce a measure of non-triviality via a contrast between the prediction error of the learned model with a naive model of the overall signal statistics. The transform can learn to discard uninformative and unpredictable components of the signal in favor of the features which are both highly predictive and highly predictable. This creates a coarse-grained model of the time-series dynamics, focusing on predicting the slowly varying latent parameters which control the statistics of the time-series, rather than predicting the fast details directly. The result is a semi-supervised algorithm which is capable of extracting latent parameters, segmenting sections of time-series with differing statistics, and building a higher-level representation of the underlying dynamics from unlabeled data.
\end{abstract}

\maketitle 

\section{Introduction}

How do physicists do feature engineering? In statistical physics, the corresponding concept for a 'golden feature' is that of an order parameter --- a single variable which captures the emergent large-scale dynamics of the physical system while projecting out all of the internal fluctuations and microscopic structures. Descriptions based entirely on a system's order parameters tend to be much more generalizable and transferable than detailed microscopic models, and can capture the behavior of many disparate systems which share some underlying structure or symmetry. The process of extracting out the large-scale dynamics of the system and discarding the microscopic details that are irrelevant to those overarching dynamics is referred to as 'coarse-graining'. In physical models, one often wants to predict dependencies between parameters or the time evolution of some variables, and while working with order-parameters means that the results become much more general, it also means that there are certain questions which become unanswerable because the details that the question depends on have coarse-grained away.

This trade-off goes hand in hand with the ability to find order parameters --- the intuition is that as one zooms out to bigger and bigger scales (and, for a physical system, anything that we interact with at the human level of experience is extremely zoomed out compared to the atomic scale), certain kinds of mistakes or mismatches between microscopic details of a model and reality will be erased, while other ones will remain relevant no matter how far you zoom out. The order parameters are then the things that are left when you have discarded everything that can be efficiently approximated as you make the system bigger. But to perform this in a self-consistent way, one must ask only for those things that matter to the large-scale details, not for anything that could be associated with some kind of error (because many errors can be defined, but not all errors will remain relevant at large scales).

If we compare this to the way in which many problems in machine learning are phrased, there is a novel element here. Usually, a loss function is designed with a specific problem in mind, and so errors in the performance of that problem are de-facto important. But if we wish to construct an unsupervised technique, it should somehow decide on its own in a way inspired on the dependencies within the data itself what is asymptotically important and what errors are irrelevant. For example, recent advances in image synthesis such as style transfer use error functions constructed out of intermediate layer activations of an object classifier network rather than working at the pixel level, with the result of minimizing perceptually meaningful inconsistencies rather than errors in the raw pixel values \cite{gatys2015neural,johnson2016perceptual}. 

So, how do you find a good order parameter? The style transfer algorithm effective uses a supervised component in order to determine what is and is not meaningful --- that is to say, the object classification task which provided the aesthetic sense to evaluate artistic styles (even though the supervised task is not directly related). Even though the supervised task didn't have to be directly related to the goal of style transfer, the details aren't irrelevant --- using a network trained to identify the source of a video frame or an autoencoder instead of an object classifier would emphasize certain aspects of the data over others. Labels about object type and location tend to emphasize edges, whereas delocalized information such as the source video clip of a frame tends to emphasize distributions of color and intensity over specific shape. Recent work has suggested that it is possible to combine generative adversarial networks and autoencoders to allow the auto-encoder to effectively discover its own loss function \cite{larsen2015autoencoding}. Here, an intrinsic predictability is used to drive the network to organize itself around the data --- specifically, the ability to predict whether something is or is not a member of the same distribution as the given data.

If we want to do this in an unsupervised fashion, we need a sense of intrinsic meaningfulness of some features over others, using only the data itself as the generator of that meaning. Since we are considering an approach in which it is permissible to declare some aspects of the data irrelevant, this becomes doubly tricky. One thing we can still do however is to require that the things we retain should be as self-predictive as possible. This brings us back to the physics analogy --- we can ask for degrees of freedom taken from point in time which then let us best predict the future of the data. This kind of approach has been used to construct things like word2vec \cite{mikolov2013distributed} to generate latent conceptual spaces for words. However, following the analogy from physics, there is a suggestion that perhaps this is asking for too much: that is to say, we are trying to predict the future microscopic variables from a set of macroscopic measurements, which may mean that we retain information solely for the purpose of spanning the microscopic basis, and not because it inherently abstracts and compresses the underlying processes which generate the data.

For example, if we were to train a word2vec on a database containing many different distinct dialects, it would be useful for predicting the 'micro' future of a sentence to know which dialect that sentence belongs to. But if we wished to model the conceptual structure of sentences, this dialect information would end up being mostly irrelevant and would force us to learn many parallel models of the same relationships, much in the way that a dense neural network has to repeatedly learn the same relationships at every pixel offset whereas a convolutional network can kernels which generalize in a translationally invariant fashion. 

\begin{figure}
\includegraphics[width=\columnwidth]{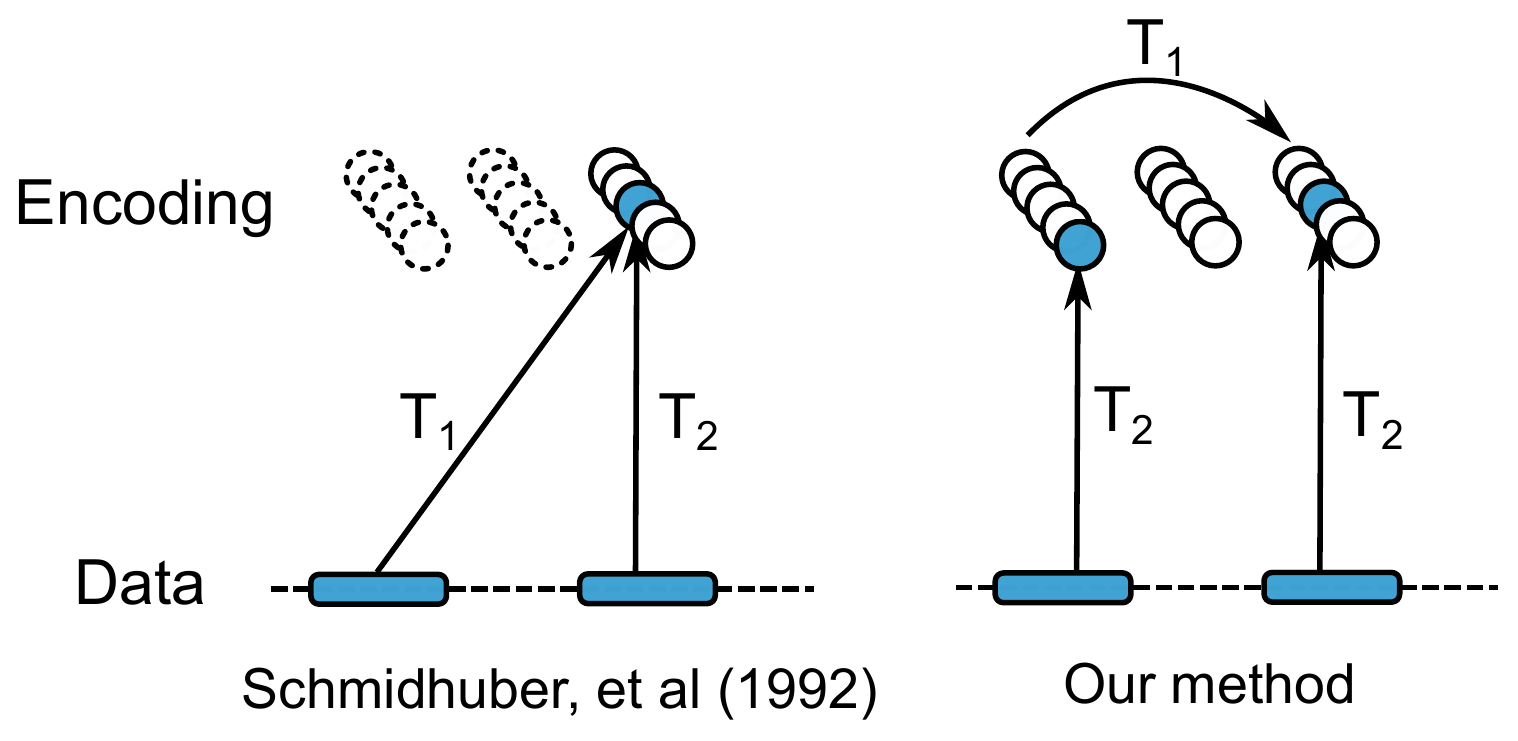}
\caption{\label{OrderParameterSchematic}Relationship between the raw data and the extracted features in Schmidhuber et al.\cite{schmidhuber1993discovering}, versus our algorithm. }
\end{figure}

This suggests that we may be able to better find good features for data to describe itself if we specifically ask for the proposed features to predict themselves, not everything about the data. Doing this allows the learner to in some sense choose its own problem to solve, finding those things which can be efficiently predicted and discarding highly unpredictable information from consideration. Work by Schmidhuber, et al.\cite{schmidhuber1993discovering} explored this idea by asking two networks to make a prediction given different views of the data --- not requiring them to make a 'correct' prediction, but only to make the same prediction --- and found that this would organize the networks to discover systematic features of the data. We extend this a bit further and ask for the new representation to contain sufficient information on its own to predict relationships and variations of the data in that representation, without specific reference to the underlying data (Fig.~\ref{OrderParameterSchematic}). What follows is a presentation of a specific algorithm and loss function able to perform this task in a stable fashion, which we will refer to as 'neural coarse-graining' or 'NCG'.

\section{Model}

\subsection{Loss function}
The key point we use in the analogy to order parameters is that an order-parameter should be self-predictive. That is to say, the microscopic model contains enough information to predict the future microscopic state, so the macroscopic model should retain just enough information to predict its own future macroscopic state (but need not predict the future microscopic state). We can think of this as two separate tasks: one task is to transform the data into a new set of variables, and the second task is to use those new variables at one point to predict the value of the new variables at a different point. The novel element is that the prediction is not evaluated with respect to matching the raw data, but is evaluated with respect to matching the transformed data --- that is to say, the network is helping to define its own loss function (in a restricted way). If a very slowly varying variable can be found, that would be favored as prediction (on shorter timescales) becomes trivial. 

This introduces a potential problem --- what if the network just projects all of the data to a constant value? In that case, the predictor would be perfect, but obviously wouldn't capture anything about the underlying data. To avoid this, we need the loss function to not just evaluate the quality of the predictions, but also somehow evaluate how hard the task was that the network set for itself. For this we take inspiration from information theory and ask, how much information is gained about the future by making a prediction contingent upon the past (relative to the stationary statistics of the signal). If we have a globally optimal predictor, then this quantity is known as the predictive information\cite{bialek1999predictive}, and is defined as:
$$\I_{\pred} = \HS(X_{\future}) - \HS(X_{\future}|X_{\past})$$
where $X_{\future}$ represents data from a signal $(X_t)_{t\in \mathbb{Z}}$ that will be observed in the future and $X_{\past}$ represents data already observed in the past. $\HS$ is the Shannon entropy. For a Markov chain the predictive information reduces to: 
$$\I_{\pred} = \HS(X_{t+1}) - \HS(X_{t+1}|X_{t})$$

In our case, we consider a transformed signal $Y_t = f(X_t)$ rather than the original signal, and want to optimize that transform to maximize the predictive information of the transformed signal. Since the transform is deterministic this predictive information turns out to be the measure of non-trivial informational closure ($\NTIC$) proposed by \citet{bertschinger_information_2006} (for a derivation see Appendix NTIC):
$$\NTIC = \HS(Y_{t+1}) - \HS(Y_{t+1}|Y_t)$$
Because the amount of information gained by conditioning on the past is bounded by the entropy of the signal being predicted, if the transformation maps to a very simple distribution then there will not be much additional information gained by knowledge of the past even if the predictor happens to be very accurate, so this protects against the projection onto a constant value. If we only wanted to construct a coarse-grained process which is predictive of its own future and captures as much information as possible about the underlying process then we could try to optimize $f$ such that $\NTIC$ is maximized.

However, we also want to adapt the coarse-graining $f$ to the capabilities of a specific (neural) predictor $g$ which given $y_t$ predicts $y_{t+1}$. In that case it can be beneficial for the transform $f$ to throw out information about $X_t$ which in general could be used to increase $\NTIC$. The information that $f$ extracts from $X_t$ should then be just the information that $g$ can predict well. The measure of $\NTIC$ does not account for such an adaptation to $g$. Nonetheless this can be done by comparing the value predicted by $g$ given $y_t$ to the value $y_{t+1}=f(x_{t+1})$ and optimizing for the accuracy of this prediction as well as for the capturing of information about $(X_t)_{t\in \mathbb{Z}}$.

We note that optimizing both $f$ and $g$ by evaluating the accuracy of the prediction by $g$ of the actual value of $y_{t+1}$ is a special case of the state space compression framework developed by \citet{wolpert_framework_2014}. 

The nature of the special case here however requires a combination of such an accuracy measure with the information extracting principle of $\NTIC$ (see also \footnote{In the notation of \citep{wolpert_framework_2014} that paper let $Y=Y$, $\mathcal{O}=\pi=f$, $\phi=g$, and $\rho=Id$. 
Then the accuracy costs they propose (e.g.\ in their Eqs.\ (4) and (11)) are similar to our term for prediction accuracy. However because the observable $\mathcal{O}$ we want to learn about here is the same as the map $\pi$ (our $f$) to the macroscopic state that we are optimizing this accuracy cost is not enough. In contrast to the usual application of the state space compression framework where $\mathcal{O}$ is externally determined here the optimization can just choose $\pi$ to map every state to a constant. This would always result in high accuracy. For this reason our accuracy cost has an additional entropy term similar to the first term in $\NTIC$.}).

While $\NTIC$ and the state space compression framework provide intuitions for our optimization function the implementation employs certain practical tweaks. The main difference to the previous discussion is that instead of optimizing a function $f$ which maps $X_t$ to random variables $Y_t$ we construct the transforms $T_2$ such that $T_2(x_t)=s_t$ can be (and is) directly interpreted as probability distribution over a macroscopic random variable $Y_t$ that does not play an explicit role itself. In other words, we treat $s_t:=T_2(x_t)$ as a probability distribution $p(Y_t|x_t)$(see note \footnote{Here we use the notation $p(A)$ with a capital $A$ to indicate the entire distribution $p:\mathcal{A}\rightarrow [0,1]$ with $\sum_{a \in \mathcal{A}} p(a) =1$ where $\mathcal{A}$ is the set of possible values of random variable $A$.}) over classes of the (implicit) classifier $Y$. This means $s_t$ has as many components as there are classes in the classifier $Y$. If $y$ denotes such a class then $s_t(y)=T_2(x_t)(y)$ is interpreted as $p(y|x_t)$. In the same way instead of optimizing a specific prediction $g$ we look at the neural transform $T_1(s_t)=\hat{s}_{t+1}$ as a probability distribution $p(\hat{Y}_{t+1}|s_t)$. All this allows us to keep the loss function smoothly differentiable with respect to changes in the transformation. 

With this in mind we optimize $T_1$ and $T_2$ to minimize:
\begin{align*}
Q \equiv - \HS(\langle s_t \rangle_t) + \langle \sum_{y \in Y} s_t(y) \log \hat{s}_t(y) \rangle_t \end{align*}

where $\langle ... \rangle_t$ indicates an average over the dataset (for example a time series indexed by $t$). This loss function combines both the optimization of the predictor $T_1$ (in the form of minimizing the cross-entropy between the true and predicted distribution in the second term) and the average entropy of the transformed signal. 

The reason for using the entropy of the dataset average of the $s_t$ is that we don't want $T_2$ to necessarily map the various data points $x_t$ to maximum entropy distributions. Each $x_t$ may well be mapped to a delta distribution-like $s_t$, instead we want $T_2$ to capture as much variation from the data as possible \textit{over time}. 

The cross-entropy on the other hand should be small at every point in time which is why the time average is taken over the instantaneous cross-entropies. The cross-entropy term takes the role of the conditional entropy term in $\NTIC$. Instead of minimizing the rest-entropy of $Y_{t}$ given $Y_{t-1}$ we here minimize the difference between the actually predicted distribution $\hat{s}_t$ and the observed distribution $s_t$. 

However, the cross-entropy does even more than that. Note that we can rewrite the cross-entropy:
$$\sum_{y \in Y} s_t(y) \log \hat{s}_t(y) = \HS(s_t) + \KL[s_t||\hat{s}_t]$$
where $\KL$ is the Kullback-Leibler divergence \footnote{The Kullback-Leibler divergence\citep[see e.g.][]{cover_elements_2006} between two distribution $p,q$ is defined as $\KL[p||q]:=\sum_x p(x) \log \frac{p(x)}{q(x)}$.}. 

Since $\KL$-divergence measures the difference between $\hat{s}_t$ and $s_t$ the entropy term $\HS(s_t)$ might seem superfluous. However, it stops $s_t$ from becoming a uniform distribution. If $T_2$ would map every $x_t$ to the uniform distribution then a loss function that omits this term (keeping only the $\KL$-divergence from the cross-entropy term) would be minimized. Due to the entropy term $T_2$ is forced to map $x_t$ to low entropy distributions which then, in contrast to uniform distributions, contain information about the particular $x_t$.

The loss function $Q$ can also generalize to partitions of the data other than past and future - any sort of partition could be used, so long as the same transformation can be applied to both sides of the partition. For example, rather than predicting the future of a timeseries, one can predict a far-away part of an image given only a local neighborhood.

The outcome of optimizing against this loss function is that the transform will extract some variable sub-component of the data that the algorithm can be very confident about, and to throw out the rest of the information in the data. By increasing the number of abstract classes or the dimensionality of the regression, this forces the algorithm to include more of the data's structure in order to maximize the entropy of the transformed data. Similar considerations govern choosing this number as would apply to choosing the size of an auto-encoder's bottleneck layer. However, the ordering of learning is opposite --- an autoencoder will start noisy and simplify until the representation fits the bottleneck, whereas this will tend to start simple and then elaborate as it discovers more things to 'say' about the data. Some care must be taken with large numbers of classes, as softmax activations experience an increasingly strong tendency to get stuck on the mean as the number of classes increases. Methods such as hierarchical softmax \cite{mikolov2013efficient} or adjusting the softmax temperature over the course of training may be useful to avoid these problems. 

\subsection{Architecture}
In order to construct a concrete implementation of neural coarse-graining, the components we need are a parameterized transform $T_2$ from the raw data into the order parameters (the coarse-graining part of the network), and a predictor $T_1$ which uses part of the transformed data to predict other nearby parts. In this paper, we implement both in a single end-to-end connected neural network. The transform network takes in raw data, applies any number of hidden layers, and then has a layer with a softmax activation --- this generates the probability distribution over abstract classes which functions as our discovered order parameter. The network then forks, with one branch simply offsetting the transformed data in time (or space), and the other branch processing the (local) pattern of classes through another arbitrary set of hidden layers, finally ending in another softmax layer. In order to evaluate the quality of the predictions, the output of the final softmax layer is then compared with the offset output of the intermediate softmax layer.

The most straightforward application of NCG is to timeseries analysis, as predicting the future given the present provides the needed locality, and sequence to future-sequence means that we can transform both the inputs and the predicted outputs using the same shared transform. As such, we examine a few applications of NCG for timeseries analysis and feature engineering, and provide a reference Python implementation of NCG using Theano\cite{theano} and Lasagne\cite{lasagne} at \url{https://github.com/arayabrain/neural-coarse-graining/}.

\section{Timeseries Analysis - Noise Segmentation}

In many cases, the observable data are being generated by indirectly by some sort of complex process governed by a small set of slowly-varying control parameters. We can use neural coarse-graining to attempt to discover these latent control parameters automatically. For example, if one had a noise signal where the detailed statistics of the noise were being slowly varied in the background but the mean of the noise remained constant, a direct attempt to auto-encode that signal to extract out the hidden feature would have difficulty as the auto-encoder would have to capture the high entropy of the noise signal before being able to accurately reproduce amplitude values. A self-predictor would be even worse, as the noise is unpredictable in detail. However, if one were to first make a new feature which described the high-order statistics of the noise in a local window, then the dynamical behavior of that feature might be highly predictable.

\begin{figure}
\includegraphics[width=\columnwidth]{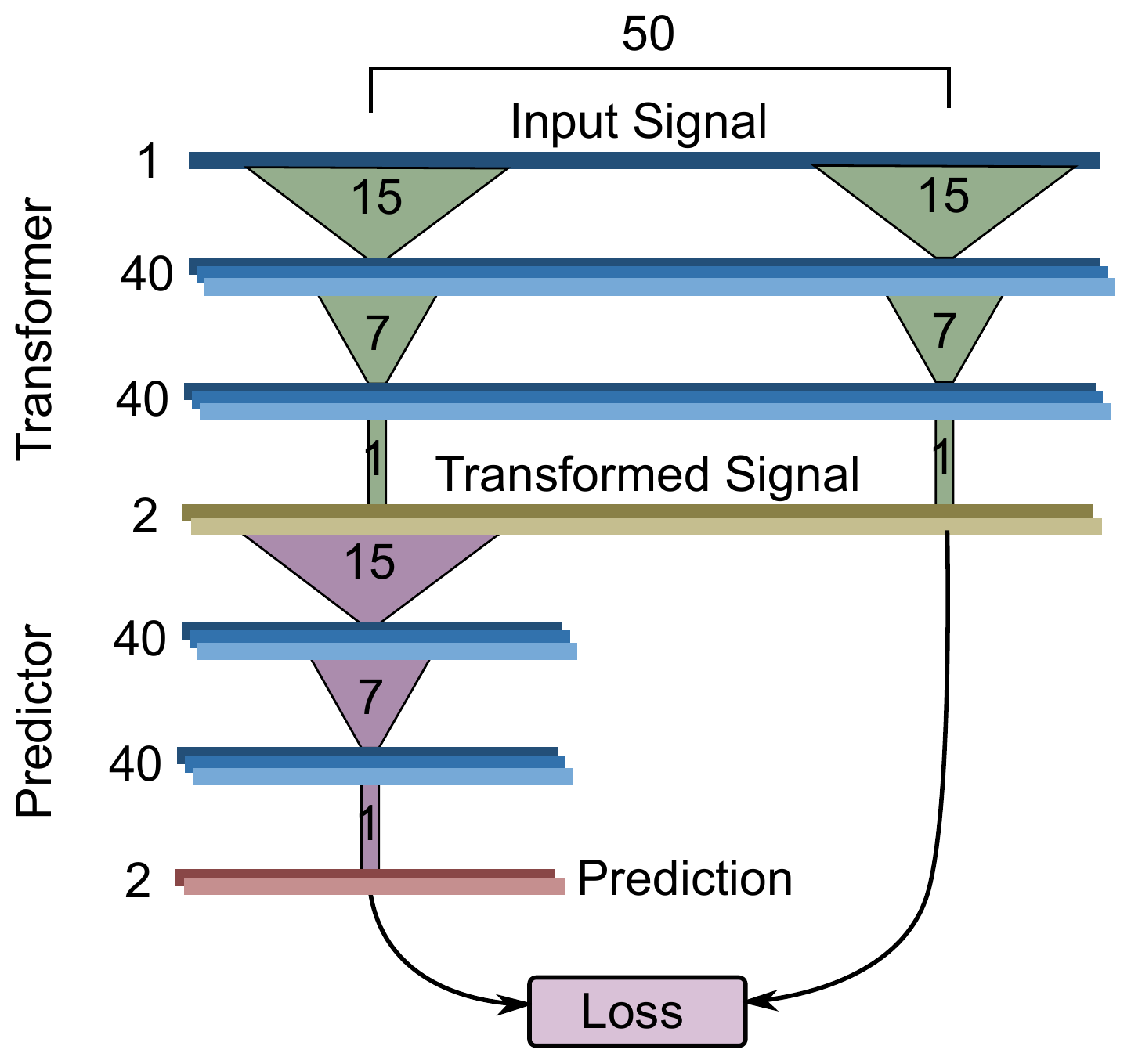}
\caption{\label{NoiseArch}Architecture used for the noise segmentation task. The two branches here do not indicate two separate networks, but rather the same convolutional operations applied at two different points in time where the offset defines the prediction timescale. When computing the loss function, prediction corresponds to matching to a time-shifted version of the transformed signal.}
\end{figure}

\begin{figure*}
\includegraphics[width=2\columnwidth]{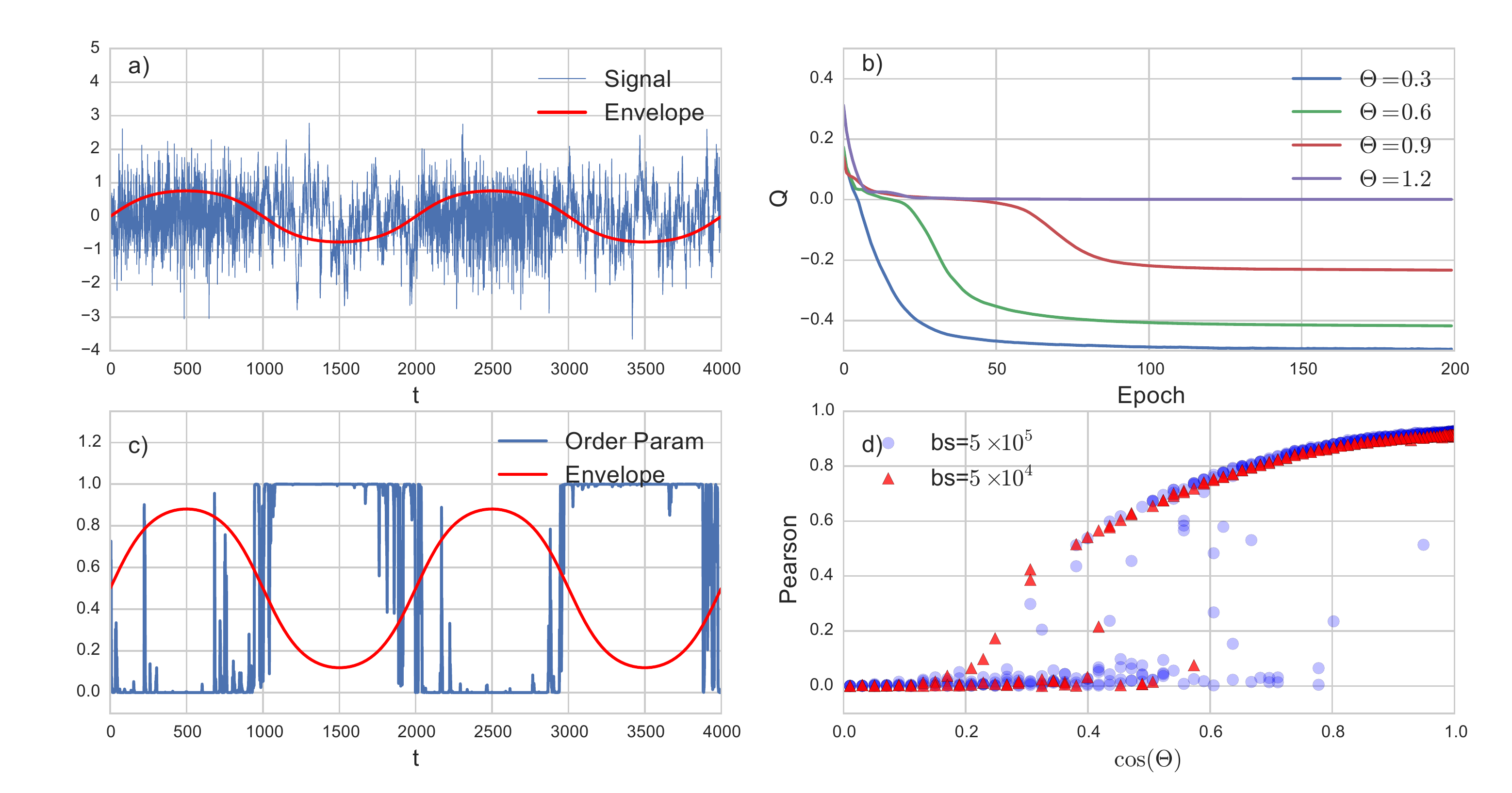}
\caption{\label{NoiseData}a) Example signal from the correlated noise segmentation task. The blue line is the raw signal, the red line is the envelope function between the two types of noise. For this example, $\cos(\Theta)=\sqrt{3}/2$. b) Training curves of the loss function for different $\Theta$ values. When the problem becomes hard, sometimes NCG gets stuck in a local minimum around the trivial prediction of assigning all transformed classes equal probability for each point in time. This trivial prediction has a loss of exactly $0$, so a plateau around $0$ is a common feature of training NCG when the problem is difficult. c) Discovered coarse-grained variable (order parameter) versus the actual envelope function for the above example. d) Pearson correlations between the discovered coarse-grained variable and the true envelope function for multiple runs at different $\Theta$ values. There is an apparent phase transition beyond which the network can no longer solve this problem and segment the noise types.}
\end{figure*}

We first consider a problem of this form in order to test the ability of neural coarse-graining to extract out the latent control parameter in an unsupervised fashion. We generate a timeseries which contains a mixture of uncorrelated Gaussian noise and auto-correlated noise, controlled by an envelope function $\psi = \frac{1}{2}(1+\tanh( sin(2\pi t/\tau) ))$, where $\tau$ controls the timescale (Fig.~\ref{NoiseData}a). We use $\tau=2000$ for these experiments. Both the independent samples and autocorrelated noise are chosen to have zero mean and unit standard deviation. To generate samples of the autocorrelated noise, we use the finite difference equation $x_t = \cos(\Theta) x_{t-1} + \sin(\Theta) \eta_t$ where $\eta$ is Gaussian noise with zero-mean and unit standard-deviation. The two noise sources are linearly combined $s_t = \psi(t)s^1_t + (1-\psi(t)) s^2_t$ to form the signal.  We generate a training set and test set of $5 \times 10^5$ samples each. An example portion of this data is shown in Fig.~\ref{NoiseData}a.

The transformer and predictor are both 1D convolutional neural networks with the joint architecture shown in Fig.~\ref{NoiseArch}. Leaky ReLU activations \cite{maas2013rectifier} with $\alpha=0.05$ are used for all layers except the output of the transformer and predictor layers, which are softmax. Batch normalization \cite{ioffe2015batch} is used on the first two layers of the transformer, and the first two layers of the predictor. The transformed signal is a probability distribution over two classes, and the predictor is attempting to predict this distribution 50 timesteps into the future (this is long enough that the receptive fields of the prediction network do not overlap in the input signal). The weights are optimized using Adam \cite{kingma2014adam}, with a learning rate of $2\times10^{-3}$, and the network is trained for $200$ epochs. Example convergence curves are shown in (Fig.~\ref{NoiseData}b). To analyze the performance of NCG with respect to discovering the envelope function we measure the Pearson correlation between the transformed signal and the known envelope function $\psi$.

When the correlation length of the auto-correlated noise is long (corresponding to small $\Theta$), NCG consistently discovers an order-parameter that is highly correlated with the envelope function (Fig.~\ref{NoiseData}c). However, as we decrease the correlation length (increasing $\Theta$), the performance drops and the different types of noise become less clearly distinguished. At a certain point, the outcome of training becomes bistable, either finding a weakly correlated order-parameter or falling into a local minimum in which the network fails to detect anything about the data. The size of this bistable region is influenced by the batch sized used in training --- if the batch size corresponds to the entire training set of $5\times10^5$, the bistable region extends from $\cos(\Theta) = [0.2,0.8]$. However, when a batch size of $5\times10^4$ samples is used, the apparent bistable region shrinks to $\cos(\Theta) = [0.2,0.6]$ (Fig.~\ref{NoiseData}d). In terms of other hyperparameters, the prediction distance and learning rate do not seem to make much difference in the ability of the network to discern between the types of noise. Increasing the size of the hidden layers likewise has very little effect. However, larger filter sizes in the transformer network do appear to have an effect, improving the Pearson correlation in large $\Theta$ cases. Even at a larger filter size, however, the bistable region appears to be unaffected.

We perform a similar test in the case of distinguishing between uncorrelated noise drawn from structurally different distributions. We compose signals which alternate between Gaussian noise and various kinds of discrete noise with the same standard deviation and zero mean. This discrete noise is taken as a generalization of the Bernoulli distribution, such that we have some number $n$ of discrete values which are selected from uniformly. We consider binary noise ($n=2$, selecting between $-1$ and $1$), balanced ternary noise (selecting from $-\sqrt{3/2},0,\sqrt{3/2}$), and unbalanced ternary noise (selecting from $-\frac{1+\sqrt{3}}{2}$, $\frac{\sqrt{3}-1}{2}$, and $1$). Whereas a linear autoregressive model can distinguish between the noise types in the previous case, the different noise distributions in this problem can only be distinguished by functions that are nonlinear in the signal variable, and so this poses a test as to whether that kind of higher-order nonlinear order parameter can be learned by the network.

In fact, this type of problem does seem to be significantly harder for NCG to solve. When the batch size is the full training set, the network appears to easily become stuck in a local minimum. As no linear features can distinguish these noise types, the network must find a promising weakly nonlinear feature before it can begin to train the predictor successfully. If the batch size is very large, the tendency is to simply decay towards a very safe uniform prediction. However, when the batch size is smaller, fluctuations are larger and have a greater chance of following a spurious linear feature far enough to discover a relevant nonlinearity. As a result, with smaller batch size the network is able to find an order parameter in the binary noise case. Furthermore, by decreasing the filter sizes (and correspondingly, putting more emphasis on deep compositions rather than wide compositions), the network becomes able to solve the unbalanced ternary noise case as well.

\begin{table}
\begin{tabular}{|cc|cc|}
 \hline
 Noise Type & Filter sizes & \multicolumn{2}{c|}{Pearson Correlation} \\
 & & Batch $5\times10^4$ & Batch $5\times10^5$ \\
 \hline
 Correlated ($\Theta=0.9$) & 15-7-1 & 0.75 & 0.76 \\
 & 25-7-1 & 0.82 & \textbf{0.84} \\
 \hline
 Binary & 15-7-1 & \textbf{0.86} & \color{red}{0.008} \\
 \hline
 Balanced Ternary & 15-7-1 & \color{red}{0.0003} & \color{red}{0.002} \\
 & 3-1-1 & \color{red}{0.003} & \color{red}{0.008} \\
 \hline
 Unbalanced Ternary & 15-7-1 & \color{red}{0.001} & \color{red}{0.0002} \\
 & 3-1-1 & \textbf{0.64} & 0.14 \\
 \hline
\end{tabular}
\caption{Results of neural coarse-graining with different filter sizes and batch sizes for the different noise segmentation test problems.}
\end{table}

\section{Timeseries Analysis - UCI HAR}

Next, we want to test whether the features generated by NCG have any practical value in other machine learning pipelines beyond just being a descriptive or exploratory tool. For this, we apply it to the problem of detecting different types of activity using accelerometer data. In this general class of problem, there is a timeseries from one or more accelerometers being worn by an individual, and the goal is to categorize what that person is doing using a few seconds of that data. The UCI Human Activity Recognition dataset \cite{anguita2013public} already contains a number of hand-designed features describing the statistics of the accelerometer data --- 2-second long chunks of the raw data are transformed into a 516-dimensional representation, taking into account measures such as the standard deviation and kurtosis of fluctuations in the raw signal. Using these engineered features, an out-of-the-box application of AdaBoost achieves 93.6\% accuracy\cite{fuhuman}.

\begin{table}
 \begin{tabular}{|c|}
 \hline
 Transformer \\
 \hline
 Conv5-100 \\
 30\% Dropout \\
 Conv3-100 \\
 30\% Dropout \\
 Conv1-20 \\
 Softmax \\
 \hline
 \end{tabular}
 \quad
 \begin{tabular}{|c|}
 \hline
 Predictor \\
 \hline
 Conv5-100 \\
 30\% Dropout \\
 Conv1-100 \\
 30\% Dropout \\
 Conv1-20 \\
 Softmax \\
 \hline
 \end{tabular}

 \caption{\label{UCINet}Architecture of the network used for UCI HAR timeseries analysis. All layers have batch normalization applied, and Leaky ReLU as the activation for the non-softmax layers.}
\end{table}

\begin{figure}
\includegraphics[width=\columnwidth]{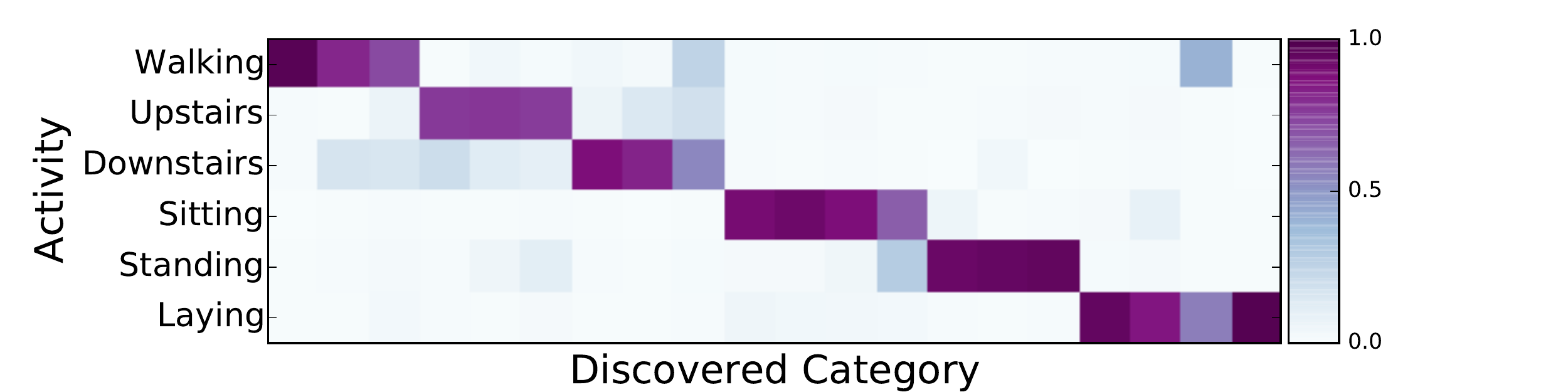}
\caption{\label{UCICorr}Correlation matrix between the discovered coarse-grained variables and the different activity classes. The columns of the matrix are sorted to bring together coarse-grained variables which are most strongly correlated with each activity in turn. Most of the coarse-grained variables are strongly associated with a single activity class, with the exceptions of columns 9, 13, and 20. }
\end{figure}

For this dataset, we transform a neighborhood of 7 timesteps of the full 516-dimensional input into 20 classes at the same temporal resolution. The prediction network takes as input a size 5 neighborhood of the transformed classes and predicts the class at 20 steps into the future. That is to say, the full receptive field of the predictor extends from $t-4$ to $t+4$ to predict the class at $t+20$ (which itself depends directly on nothing earlier than $t+17$). The network uses leaky ReLU with $\alpha=0.05$, batch normalization on each layer, and 30\% dropout between each layer. A schematic of the full architecture is given in Table~\ref{UCINet}. The data is split into chunks of length 120 steps, and the network is trained for 510 epochs using Adam optimization with a learning rate of $5\times10^{-3}$.

The resulting classes already show strong correlations with the different behaviors (Fig.~\ref{UCICorr}). However, it's clear that for some samples the discovered classes are ambiguous and do not uniquely identify the behavior. When we use these new features from three adjacent timesteps with AdaBoost (in the form of Scikit-Learn's GradientBoostingClassifier implementation \cite{scikit-learn}) on their own, we observe only 87.4\% accuracy on the test set, compared with 93.7\% using the original features. However, when we combine the original features with our classes, the accuracy on the test set increases to 95.2\%. So the new features seem to expose some structures in the data which are otherwise more difficult to extract by AdaBoost itself.

One confounding factor here is that our algorithm had access to multiple timesteps, whereas the only temporal information available to the original score was from the 2-second interval used to produce the hand-designed features in the original data. As such, it may be that this increase in performance is only due to the availability of a wider time window of inputs. We test this by measuring the performance using the original features, but taken at $t-8$, $t$, and $t+8$ in order to approximate the range of access that our algorithm was provided. This improves the performance as well, resulting in 94.6\% accuracy on the test set. However, when we take the time-extended original features and combine with our discovered features, the performance is worse than just using the instantaneous original features with our classes (95.1\%). This seems to suggest that some degree of what the discovered order parameters are doing is to efficiently summarize coherent aspects of the time-dependence of the data.

\begin{table}
\begin{tabular}{|c|c|}
 \hline
 Features & Accuracy \\
 \hline
 Original (1 fr.) & 93.7 \\
 Original (3 fr.) & 94.6 \\
 NCG (3 fr.) & 87.4 \\
 \textbf{NCG + Original (1 fr.)} & \textbf{95.6} \\
 NCG + Original (3 fr.) & 95.1 \\
 \hline
\end{tabular}
\caption{\label{UCIResults} Results of AdaBoost classifier on the UCI HAR dataset trained with different sets of features --- the original features from one or three frames, and the neural coarse-graining features with one or three frames (and in combination with the original features). The NCG features produce a worse classifier on their own, but result in an improvement over the original features alone when combined.}
\end{table}

\begin{figure}
\includegraphics[width=\columnwidth]{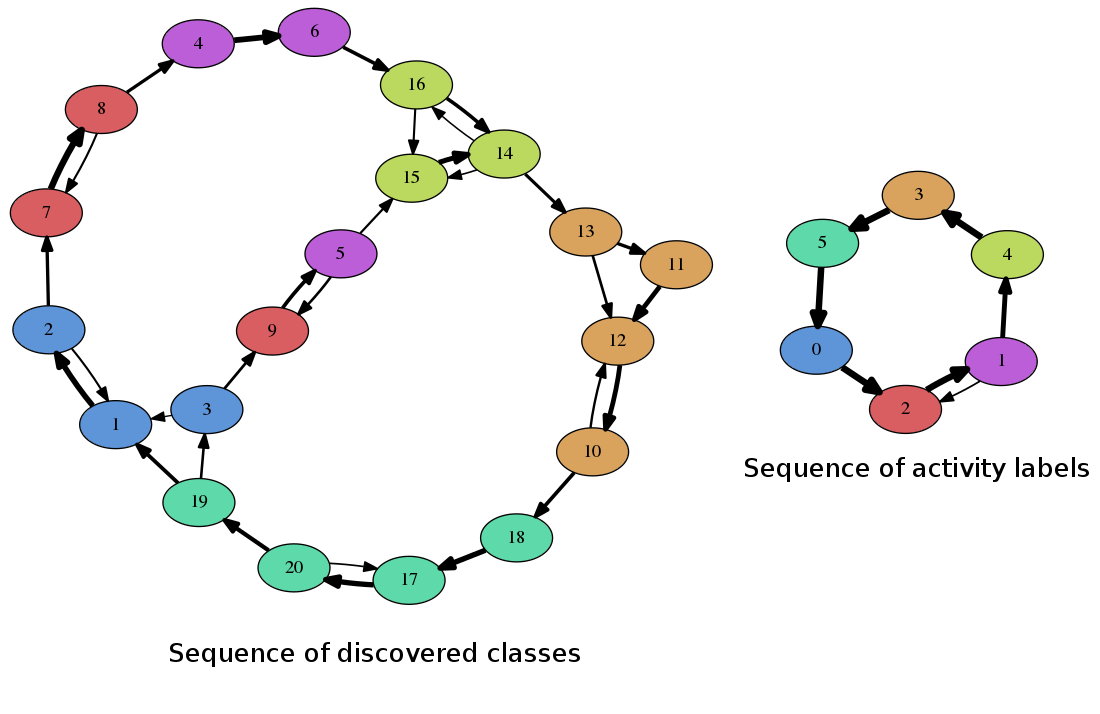}
\caption{\label{Transitions}Left: Graph of the transitions between discovered categories in the transformed data. Links are drawn for transitions which occur with over a 20\% probability from the previous class. The numbering of nodes in this plot corresponds to the column order in Fig.~\ref{UCICorr}. Node colors are based on the corresponding highly-correlated activity. Right: Graph of the transitions between activities in the raw data. The activities were always performed in a fixed order, so this cyclic structure ends up strongly determining the behavior of long-term temporal sequence predictions --- possibly an artefact that NCG is picking up on in generating its features for this problem.}
\end{figure}

We can also use the classes generated by neural coarse-graining to do exploratory analyses of the data. Since the transformed data tends to be locally stable with sharp transitions between the categories, a natural thing to do is to sample the between-class transitions as a Markov process. For the UCI HAR dataset there's a complication --- the data was taken according to a specific protocol, ordering the activities to be performed in a specific way for each subject. The length of time spent on each activity is quite short, so it would be hard to make predictions that did not cross some activity boundary. It turns out that our algorithm ends up predominantly discovering this structure in the resulting Markov process (Fig.~\ref{Transitions}). The regularity of the protocol means that discovering what activity is currently under way is a very good predictor of what activity will be taking place in the future, and as such probably strongly encouraged the order parameters to be correlated with the activity types. The double-loop structure may indicate the discovery of some subject-specific details.

\section{Conclusions}

We have introduced the neural coarse-graining algorithm, which extracts coarse-grained features from a set of data which are both readily determined from the local details, and which also are highly predictive of themselves. The coarse-graining does not preserve all underlying relationships from the data, but instead tries to find some subset of those relationships which it can most readily predict. This provides a form of unsupervised labelling, to map a problem onto a simpler sub-problem that discards the parts of the data which confound prediction. One advantage of this approach compared to directly using self-prediction on the underlying data is that neural coarse-graining is free to predict parameters controlling the distribution of the noise rather than the details of the individual random samples, making this method more robust to prediction tasks on highly noisy data sets, including ones where the structure of the noise may be important to understand or take into account.

Although neural coarse-graining trains a predictor, it is unclear whether in general the predictor itself is useful towards any particular task. Rather, it is the way in which needing to be able to construct a predictor forces the transformation to preserve certain features of the data over others. As such, the sub-problem that the network decides to solve can be used for exploratory analysis to characterize the dominant features of the underlying processes behind a set of data. By examining the transition matrix between discovered classes, it is possible to extract a coarse-grained picture of the dynamics, detecting things such as underlying periodicities or branching decision points in time-series data. In addition, from our experiments on UCI HAR, it seems that the extracted features may capture or clean up details of the underlying data in a way that can be used to augment the performance of other machine learning algorithms --- a form of unsupervised feature engineering.

\section{Acknowledgments}
Martin Biehl would like to thank Nathaniel Virgo and the ELSI Origins Network (EON) at the Earth-Life-Science Institute (ELSI) at Tokyo Institute of Technology for inviting him as a short-term visitor. Part of this research was performed during that stay.

\appendix*
\section{Appendix: NTIC}
\label{ntic}
We show that optimizing non-trivial informational closure reduces to optimizing $\I_{\pred}$ of the transformed signal. In general if we have two processes $(X_t)_{t \in \mathbb{Z}}$ and $(Y_t)_{t \in \mathbb{Z}}$ and we assume that the joint process $(X_t,Y_t)_{t \in \mathbb{Z}}$ is a Markov chain then non-trivial informational closure of $(Y_t)_{t \in \mathbb{Z}}$ with respect to $(X_t)_{t \in \mathbb{Z}}$ is measured by (at any time $t$):
\begin{equation}
\label{eq:ntic}
 \NTIC := \I(X_t:Y_{t+1}) - \I(Y_{t+1}:X_t|Y_t).
\end{equation} 
The smaller this value the more non-trivially closed is $Y_t$. The first term measure how much information is contained in $X$ about the future of $Y$. The second term measure how much more information $X$ contains about the future of $Y$ than is already contained in the present of $Y$. The process $Y$ is called non-trivially closed with respect to $X$ because it shares information with $X$ but this information is contained in $Y$ itself.

In our case $Y_t=f(X_t)$ where $f$ is a deterministic transform. Therefore the second term in Eq. \ref{eq:ntic} reduces to 
\begin{equation}
 \I(Y_{t+1}:X_t|Y_t) = \HS(f(X_{t+1})|f(X_t)) - \HS(f(X_{t+1})|X_t).
\end{equation} 
Writing the first term also via the entropies we can easily see that 
\begin{equation}
 \NTIC=\HS(f(X_{t+1}))-\HS(f(X_{t+1})|f(X_t)).
\end{equation} 
This is just the Markov chain approximation of $\I_{\pred}$ of $(Y_t)_{t \in \mathbb{Z}}$.

\section{Appendix: Continuous order parameters}
\label{regression}
When constructing things in terms of information measures, it is easier to use discrete states rather than continuous states. However, a continuous version of the coarse-graining loss function can be constructed to extract continuous-valued variables. The interpretation of these is a bit simpler, as it doesn't require treating the transformed variable as a distribution and as a value in different parts of the algorithm. In order to make this construction, we must be able to compare the entropy of the transformed signal from a naive point of view with the entropy of the transformed signal conditioned on the predictor. To do this, we generate two signals: the signal corresponding to the coarse-grained variable $y(t)$ and the 'residual' signal corresponding to the prediction error against the future coarse-grained signal $\epsilon \equiv P(y(t+\Delta)|y(\tau<t)) - y(t+\Delta)$. We then construct a loss function which measures the difference in entropies between $y(t)$ and $\epsilon(t)$.

To do so, we must assume something about the distributions of $y$ and $\epsilon$. If we assume that these signals correspond to samples taken from a multi-dimensional Gaussian distribution, then the entropy of each signal corresponds to the logarithm of the determinant of the covariance matrix. This lets us construct a regression loss function for continuous coarse-grained variables.

$$ Q_{reg} \equiv \log( \frac{\det \textrm{cov}(y,y)}{\det \textrm{cov}(\epsilon, \epsilon)} ), $$

We mostly include this example for completeness and as a demonstration of how to construct coarse-graining loss functions for different types of variable. Although this form may be conceptually more tidy than the discrete case, we have found that in general the discrete version of this algorithm seems to perform better and is less prone to overfitting, at least in those cases which we have investigated. 

\nocite{cover_elements_2006}
\bibliography{references}

\end{document}